\PassOptionsToPackage{table}{xcolor}
\documentclass[twocolumn]{galbot}

\definecolor{draftcolor}{HTML}{8A5100}

\colorlet{tabheadcolor}{galbot!20}
\colorlet{tabstripecolor}{galbotlight}
\colorlet{taburscolor}{galbotnavy!10}
\newcommand{\tabheader}{\rowcolor{tabheadcolor}}
\newcommand{\tabours}{\rowcolor{taburscolor}}
\newcommand{\galbottablestyle}{%
  \rowcolors{1}{white}{tabstripecolor}%
  \arrayrulecolor{galbot!75}%
  \setlength{\aboverulesep}{0pt}%
  \setlength{\belowrulesep}{0pt}%
  \renewcommand{\arraystretch}{1.18}}
\AtBeginEnvironment{table}{\galbottablestyle}
\AtBeginEnvironment{table*}{\galbottablestyle}

\usepackage{wrapfig}
\usepackage{tabularx}
\usepackage{textcomp}
\usepackage{stfloats}
\usepackage{url}
\usepackage{verbatim}
\usepackage{graphicx}
\usepackage{titlesec}
\usepackage{tocloft}
\usepackage{adjustbox}
\usepackage{multirow}
\usepackage{tikz}
\usepackage{comment}
\usepackage{amsmath,amssymb}
\usepackage{colortbl}
\usepackage{color}
\usepackage{booktabs} 
\usepackage{hyperref}
\usepackage{subcaption} 
\usepackage{arydshln}
\RequirePackage{xspace}
\makeatletter
\DeclareRobustCommand\onedot{\futurelet\@let@token\@onedot}
\def\@onedot{\ifx\@let@token.\else.\null\fi\xspace}

\def\vs{\emph{vs}\onedot}

\makeatother

\usepackage{makecell}

\usepackage{pifont}
\usepackage{bbding}
\usepackage{fontawesome}
\usepackage{xspace}

\usepackage{float}

\newlength\savewidth

\newcolumntype{x}[1]{>{\centering\arraybackslash}p{#1pt}}
\newcolumntype{y}[1]{>{\raggedright\arraybackslash}p{#1pt}}
\newcolumntype{z}[1]{>{\raggedleft\arraybackslash}p{#1pt}}

\renewcommand{\paragraph}[1]{\vspace{1mm}\noindent\textbf{#1}}

\usepackage{xcolor}
\usepackage{array}
\usepackage{bbm}
\usepackage{collcell,xfp}
\usepackage{pgf}
\usepackage[most]{tcolorbox}
\usepackage{csquotes}
\usepackage[noorphans,vskip=1em,leftmargin=1em]{quoting}
\usepackage{enumitem} 
\usepackage{forest}
\usepackage{caption}
\usepackage{longtable}
\ifPDFTeX
  \usepackage[T1]{fontenc}
\fi

\renewcommand{\paragraph}[1]{\vspace{1.25mm}\noindent\textbf{#1}}

\usepackage{algorithm}
\usepackage{listings}

\definecolor{codeblue}{rgb}{0.25, 0.5, 0.5}
\definecolor{codekw}{rgb}{0.35, 0.35, 0.75}
\lstdefinestyle{Pytorch}{
    language = Python,
    backgroundcolor = \color{white},
    basicstyle = \fontsize{9pt}{8pt}\selectfont\ttfamily\bfseries,
    columns = fullflexible,
    aboveskip=1pt,
    belowskip=1pt,
    breaklines = true,
    captionpos = b,
    commentstyle = \color{codeblue},
    keywordstyle = \color{codekw},
}

\definecolor{green}{HTML}{009000}
\definecolor{red}{HTML}{ea4335}

\definecolor{linecolor1}{RGB}{246, 248, 239}
\definecolor{linecolor2}{RGB}{230, 234, 217}
\definecolor{linecolor3}{RGB}{211, 222, 190}

\newcommand{\methodname}{ScaleVLA\xspace}

\title{ScaleVLA: Scaling Vision-Language-Action Models to 552 Billion Parameters}

\author[1,4,*]{Ruixi Yu}
\author[2,4,*]{Dairu Liu}
\author[1,4,*]{Xuchuan Chen}
\author[1,4,*,\ddagger]{Zekun Qi}
\author[1,4]{Zekai Li}
\author[1,4]{Zhuo Cao}
\author[2,4]{Luyang Niu}
\author[1,4]{Yingsheng Zhang}
\author[3,4]{Sikai Liang}
\author[3,4]{Ke Fan}
\author[2,4,\dagger]{He Wang}
\author[1,4,\dagger]{Li Yi}
\affiliation[1]{Tsinghua University}
\affiliation[2]{Peking University}
\affiliation[3]{Shanghai Jiao Tong University}
\affiliation[4]{Galbot}
\contribution[*]{Equal contribution}
\contribution[\ddagger]{Project leader}
\contribution[\dagger]{Corresponding authors}

\date{\today}
\page{\url{https://galaxygeneralrobotics.github.io/scalevla/}}

\abstract{
Frontier Vision-Language Models (VLMs) such as GPT-6 Astra have demonstrated remarkable capabilities in robot control, suggesting that broad pretraining provides knowledge valuable for physical action. Yet most Vision-Language-Action (VLA) policies still rely on much smaller VLM backbones, leaving this potential largely untapped. We ask whether frontier-scale representations can provide a stronger foundation for robot learning. We introduce \methodname, which connects a 552B DeepSeek-V4.1-Flash backbone to a trainable action DiT, and systematically compare VLMs from 2B to 552B under a shared training setup. We evaluate three fundamental capabilities of robot foundation models. In \emph{sample efficiency}, just 10 demonstrations outperform a 2B backbone trained with 50 (87.8\% vs. 85.6\%). In \emph{generalization}, LIBERO-PRO success reaches 45.1\%, compared with at most 31.3\% for smaller backbones. In \emph{precise manipulation}, real-world charger insertion improves from 4/10 to 9/10 successes, with gains concentrated in contact-sensitive alignment. Despite its scale, \methodname achieves 216\,ms median latency per action chunk. Our findings suggest that frontier VLMs can serve not only as reasoning agents, but also as powerful representational foundations for efficient and generalizable robot policies.

}

\begin{document}
\maketitle
\pagestyle{plain}

\begin{figure*}[t]
  \centering
  \includegraphics[width=\textwidth]{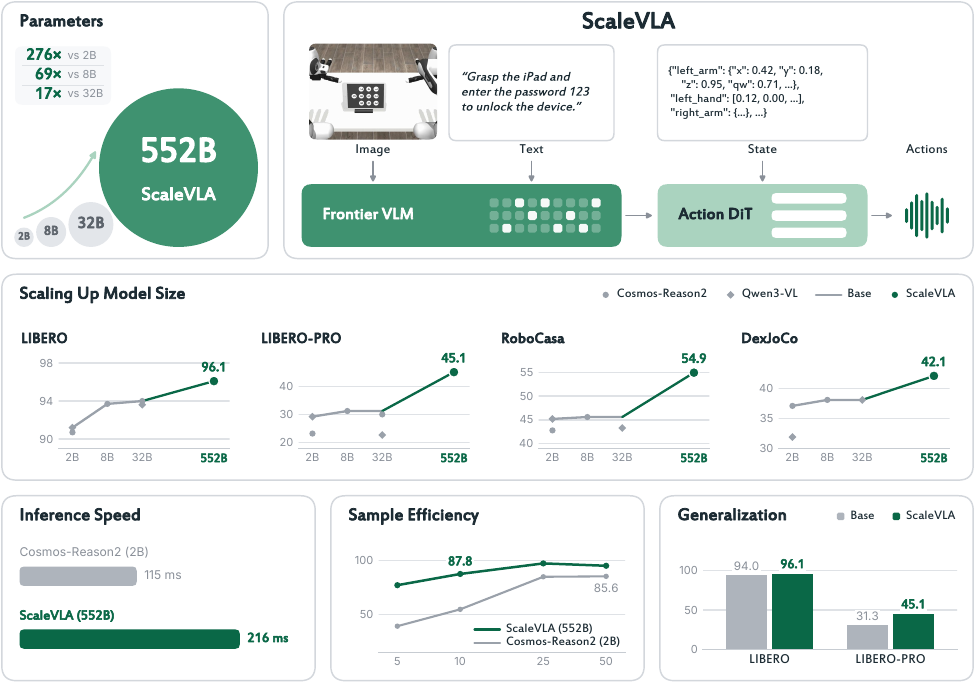}
  \caption{\textbf{\methodname overview.}
  Top left: backbone sizes. Top right: the VLM encodes images and the instruction, and the action DiT uses these features and robot state to generate an action chunk.
  Middle: task success by backbone size. Base is the best dense backbone at or below each size.
  Bottom: inference latency on RoboCasa, sample efficiency on LIBERO Goal, and success on LIBERO and LIBERO-PRO.}
  \label{fig:teaser}
\end{figure*}

\section{Introduction}
\label{sec:intro}

Frontier vision-language models are beginning to show remarkable capabilities in physical control. Recent studies of GPT-6 Astra demonstrate that a general-purpose model can generate robot actions, perform manipulation, and improve its behavior through physical feedback~\cite{galbot2026astra,astranomy2026,li2026physrsi}. What makes these results particularly intriguing is that such capabilities are not acquired primarily from robot demonstrations. Instead, they appear to draw on the broad knowledge developed during large-scale pretraining, including visual perception, spatial understanding, reasoning, and code generation~\cite{codeaspolicies2023,spatialvlm2024}. These observations suggest a compelling possibility: \emph{the knowledge needed for more capable robot policies may already exist within frontier VLMs.}


Yet most vision-language-action models still rely on relatively small VLM backbones, such as the 2B Cosmos-Reason2 in GR00T N1.7~\cite{gr00tn17}. While effective, these models may not fully capture the rich knowledge emerging in frontier VLMs.
This raises a fundamental question: \emph{can frontier-scale representations unlock greater potential in learned robot policies?}

To investigate this question, we introduce \methodname, which scales the VLM backbone of a learned robot policy to 552 billion parameters using DeepSeek-V4.1-Flash-1017~\cite{deepseekai2026deepseekv41flash}. Inspired by the action-expert architecture of GR00T~\cite{gr00tn1}, we connect a VLM to a trainable diffusion transformer (DiT) that converts multimodal representations into continuous robot actions. Rather than relying on explicit reasoning at every control step, the policy learns to directly translate pretrained representations into actions. We systematically compare VLM backbones ranging from 2B to 552B parameters, holding the action architecture, initialization, robot demonstrations, and optimization budget fixed. This allows us to examine how much robot learning can benefit from stronger pretrained representations without introducing additional action-side supervision or pretrained robot-action experts.

We argue that the value of a robot foundation model should be measured not only by its performance on familiar tasks, but by \emph{how efficiently it learns, how well it generalizes, and how precisely it acts}. We therefore evaluate \methodname\ across LIBERO~\cite{libero}, LIBERO-PRO, RoboCasa~\cite{robocasa}, DexJoCo~\cite{dexjoco}, and real-world manipulation, focusing on three fundamental capabilities.
\textbf{First, sample efficiency:} on LIBERO-Goal, \methodname\ achieves 87.8\% success with only 10 demonstrations per task, surpassing a 2B backbone trained with 50 demonstrations (85.6\%).
\textbf{Second, generalization:} under LIBERO-PRO perturbations, it achieves 45.1\% average success, compared with at most 31.3\% for the smaller backbones, indicating stronger transfer beyond the training distribution.
\textbf{Third, precise manipulation:} on a real bimanual charger-insertion task, it succeeds in 9 out of 10 reference-layout trials, compared with 4 out of 10 for the 2B backbone. The improvement is concentrated in contact-sensitive alignment and insertion rather than reaching or grasping. Together, these results suggest that frontier representations are particularly valuable when robot data is scarce, the environment changes, or fine-grained coordination is required.

Crucially, frontier-scale capacity does not necessarily imply prohibitive inference latency. DeepSeek-V4.1-Flash uses a sparse mixture-of-experts architecture, activating only a fraction of its 552B parameters per token~\cite{deepseekmoe2024,deepseekai2026deepseekv41flash}. Moreover, \methodname\ extracts features from the causal encoder of the VLM without autoregressive reasoning or text generation. It achieves a median inference latency of 216\,ms per action chunk using four H200 GPUs (\cref{tab:latency_robocasa}). This demonstrates that frontier VLM representations can support practical, subsecond robot control, although at a higher hardware cost than smaller backbones.
On the real robot, lower latency does not translate into faster execution: although \methodname has the highest per-chunk latency of the three policies, it completes successful trials in a median of 114.4\,s, 13\% less time than $\pi_{0.5}$ + Action Head (132.0\,s), because it aligns and inserts the plug 28\% faster (\cref{tab:real}).

Our findings point toward a promising direction for VLA development. Frontier pretrained representations can substantially improve data efficiency, robustness, and contact-sensitive manipulation. Rather than treating frontier VLMs only as high-level reasoning agents, we show that they can serve as the representational foundation of fast, learned robot policies. We hope this work encourages a shift toward fully exploiting frontier-model pretraining as a central ingredient in building more capable and generalizable VLAs.

Our contributions are threefold:
\begin{itemize}[leftmargin=*,nosep]
    \item We introduce \methodname, a VLA framework that connects a 552B frontier VLM to a trainable action DiT, enabling continuous robot control without autoregressive reasoning.
    \item We systematically investigate backbone scaling from 2B to 552B across simulation and real-world manipulation, demonstrating substantial gains in sample efficiency, generalization, and contact-sensitive control.
    \item We demonstrate practical subsecond inference with a sparse frontier VLM, showing that large pretrained representational capacity can be brought into learned robot policies at manageable latency.
\end{itemize}

\section{Related Work}
\label{sec:related_work}

\paragraph{VLM as Policy.}
Language models select skills or write task plans~\cite{saycan2022,innermonologue,progprompt}. PaLM-E conditions planning on sensor observations~\cite{palme}. Other systems generate controller code, spatial value maps, or keypoint constraints for motion planning~\cite{codeaspolicies2023,instruct2act,voxposer2023,rekep}. Studies of Astra assess direct action generation, cooperation with learned policies, and improvement through feedback with fixed model weights~\cite{galbot2026astra,astranomy2026,li2026physrsi}.

\paragraph{Vision-Language-Action Models.}
Early policies learn from diverse tasks or multimodal prompts~\cite{gato,rt1,bcz,robocat,vima,languagetable}. Cross-embodiment policies extend this approach across robots~\cite{openx2023,octo2024,hpt,crossformer,xvla,univla}. BridgeData V2 and DROID expand real robot data, MimicGen synthesizes demonstrations, and EgoDex provides egocentric hand trajectories~\cite{bridgev2,droid,mimicgen,egodex}. RT-2, OpenVLA, and FAST represent actions as tokens~\cite{rt2,openvla,fast}. Continuous action chunks can be predicted by regression, diffusion, or flow matching~\cite{act,oft,dp,cogact,rdt,pi0,gr00tn1}. Recent VLAs study open-world transfer~\cite{pi05,geminirobotics} and cross-embodiment alignment~\cite{qwenrobotmanip}.

Visual pretraining supplies transferable features~\cite{r3m,mvp,vc1,theia}, including DINOv2 and SigLIP features used in OpenVLA~\cite{dinov2,siglip,openvla}. RoboFlamingo adapts pretrained VLMs for imitation, GR00T N1.5 freezes them, and Knowledge Insulation separates action-expert gradients from VLM adaptation~\cite{robotflamingo,gr00tn15,knowledgeinsulation}. Compact models and efficient action interfaces reduce training or inference cost~\cite{tinyvla,robomamba,smolvla,vlaadapter,efficientvla2026}. Controlled comparisons revisit backbone choice, feature readout, and policy design~\cite{robovlms,vlm4vla,vlmrepr2026,starvla}. Demonstration sweeps test how much robot data adaptation requires~\cite{howmanydemos2026}. We compare backbones up to 552B with a fixed action DiT and demonstration budget.

\paragraph{Spatial Reasoning for Robot Manipulation.}
Manipulation policies use language-conditioned spatial maps, multiview features, or point clouds~\cite{cliport,peract,rvt,rvt2,dp3}. Spatial supervision ranges from points and 3D references to action traces~\cite{spatialvlm2024,robopoint2024,roborefer2025,zhou2026robotracer,yu2026groundingpi,llarva}. Intermediate language, trace prompts, and high-level trajectory predictions can guide action selection~\cite{ecot,cotvla,rt_h,tracevla,hamster}.

\paragraph{World Action Models.}
Video can specify a task, supply visual subgoals, or provide synthetic demonstrations~\cite{vid2robot,physicalintelligence2026pi07,dreamgen}. UniPi recovers actions from generated video~\cite{unipi2023}. Other approaches learn visual and action predictions together~\cite{gr12023,gr2,vla3d,videovla,nvidia2026gr00t2,dreamzero,cosmospolicy,motus}. DreamVLA forecasts perceptual features, DeFI separates forward and inverse dynamics, and ImageWAM uses image-editing features for control~\cite{dreamvla2025,defi2026,imagewam2026}.

\section{Method}
\label{sec:method}

\begin{figure*}[t]
  \centering
  \includegraphics[width=\textwidth]{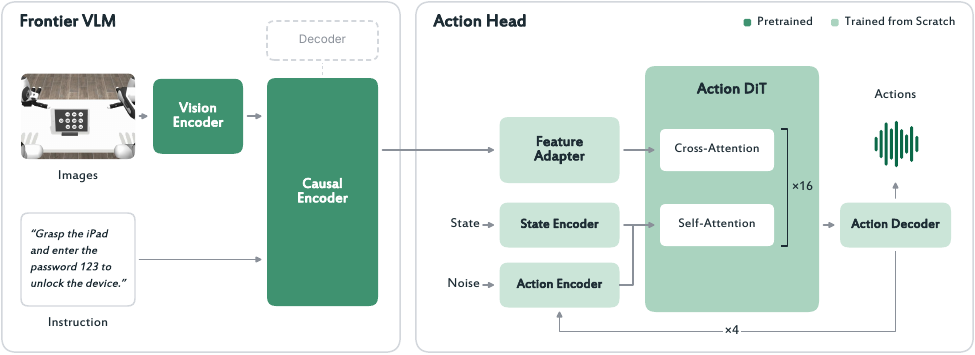}
  \caption{\textbf{\methodname architecture.}
  The VLM encodes the images and the instruction, and the action head reads the output of its causal encoder.
  The action DiT alternates cross-attention and self-attention and generates an action chunk in four flow steps.}
  \label{fig:architecture}
\end{figure*}

\subsection{Model Architecture}
\label{sec:overview}

\methodname has two parts (\cref{fig:architecture}): a frontier vision-language backbone that encodes the camera images and the language instruction, and an action head trained from scratch that turns these representations and the robot state into a chunk of continuous actions with flow matching.
Evaluations of GPT-6 Astra show useful manipulation through feedback-guided interaction and cooperation with learned policies, but repeated reasoning and action review make each decision slow and costly~\cite{galbot2026astra}.
\methodname does not reason in text: it reads the backbone once per action chunk and maps its representations directly to actions.
The backbone supplies the knowledge of frontier-scale pretraining, and the action head learns continuous control from robot demonstrations.
The language and visual backbones are frozen, while the feature adapter, state and action encoders, diffusion transformer, and action decoder are trainable.


\paragraph{Vision-language backbone.}
\label{sec:backbone}
We adopt DeepSeek-V4.1-Flash-1017~\cite{deepseekai2026deepseekv41flash} as the vision-language backbone, a mixture-of-experts model with 552B parameters.
A vision encoder converts camera images into visual tokens, and a language backbone, organized into a 20-layer causal encoder and a 20-layer decoder, processes them together with the instruction in the reference chat format of the model.
We run the causal encoder and read the features supplied to the attention operation of the next layer: a sequence of visual and text tokens, each with $D_{\mathrm{vlm}}=5120$ dimensions.
Computation stops at this interface, so the decoder never runs.
We keep every token in order, with its modality label, so the action head can attend to image and non-image tokens separately.

\paragraph{Action head.}
\label{sec:action_head}
The action head follows the GR00T N1.7 architecture~\cite{gr00tn17} and consists of a feature adapter, state and action encoders, a diffusion transformer (DiT), and an action decoder, all initialized randomly and trained on robot demonstrations.
The feature adapter applies layer normalization and a two-layer linear projection that maps each backbone token from $D_{\mathrm{vlm}}$ to a common width of 2048, followed by a four-layer self-attention transformer in which visual and language features interact under action supervision.
Embodiment-specific multilayer perceptrons encode the robot state and the noisy actions, padded to a common layout; each future action becomes a token with a learned position embedding, and the action encoder also receives the flow time.
The action DiT has 32 layers that alternate self-attention and cross-attention.
Self-attention runs jointly over the state and noisy action tokens, and cross-attention reads the adapted backbone features; successive cross-attention layers attend to non-image and image tokens in turn, and token masks exclude padding.
Adaptive layer normalization supplies the flow time to the transformer, and an embodiment-specific action decoder maps the output action tokens to a predicted velocity for the whole chunk.
Across backbone comparisons, the action head keeps the same architecture and common random initialization; only the input width $D_{\mathrm{vlm}}$ of the feature adapter follows the backbone.

\subsection{Training and Inference}
\label{sec:training_inference}

\paragraph{Flow matching.}
We train the action head with flow matching~\cite{lipman2023flowmatching}.
Given a demonstration chunk $A$, we sample Gaussian noise $\epsilon\sim\mathcal{N}(0,I)$ and a flow time $\tau\in[0,1]$, and form the noisy chunk $A^\tau=(1-\tau)\,\epsilon+\tau A$, which moves linearly from noise to the demonstration.
The action head learns the velocity $A-\epsilon$ of this path,
\begin{equation}
\mathcal{L}_{\mathrm{FM}}
=
\mathbb{E}\left[
\left\|
v_{\theta}(A^\tau,\tau\mid Z,s,e)
-
(A-\epsilon)
\right\|_{\mathrm{valid}}^2
\right],
\label{eq:flow_loss}
\end{equation}
where $Z$ denotes the adapted backbone features, $s$ the robot state, and $e$ the embodiment.
The loss averages squared error over valid action dimensions and timesteps.
We normalize state and action values using the 1st and 99th percentiles of the training data.
Gradients update only the action head, so we precompute the backbone features once and reuse them during training.

\paragraph{Action chunk generation.}
At inference, the action head starts from Gaussian noise and integrates the predicted velocity in four steps to produce an action chunk.
All four steps reuse the same backbone features, so the backbone runs once per chunk.
We undo action normalization and execute the resulting numerical commands sequentially.
\methodname predicts 16-step chunks for LIBERO, where eight commands are executed before replanning, and 30-step chunks for DexJoCo.
DexJoCo buffers timestamped commands and generates replacement chunks asynchronously while the controller executes available actions.
This separates command execution from VLM inference.
We measure execution rate and action-chunk latency separately to assess sustained control.

\section{Experiments}
\label{sec:experiments}

We compare policies with VLM backbones from 2B to 552B parameters. We measure task success in simulation, time policy inference, and vary the number of demonstrations to measure sample efficiency. We then evaluate the policies on a bimanual robot.

\subsection{Experimental Setup}
\label{sec:exp_setup}

\begin{figure*}[t]
  \centering
  \vspace{-2pt}
  \includegraphics[width=\textwidth]{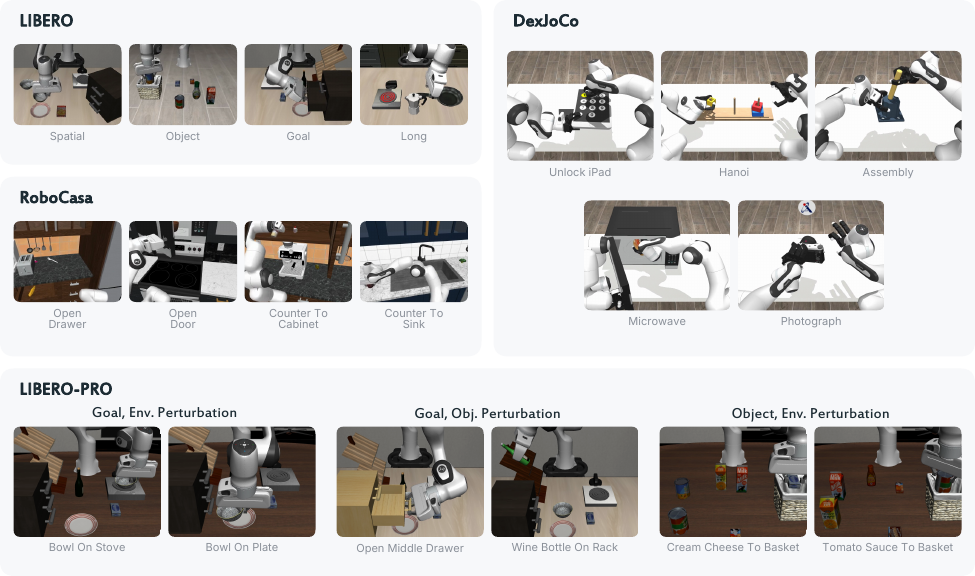}
  \caption{\textbf{Simulation tasks.}
  LIBERO tests tabletop manipulation, LIBERO-PRO perturbs the objects and scenes of LIBERO, RoboCasa tests kitchen tasks in out-of-distribution scenes, and DexJoCo tests five tasks that require coordinating two dexterous hands.}
  \label{fig:tasks}
\end{figure*}

\paragraph{Benchmarks.}
LIBERO~\cite{libero} covers language-conditioned gripper manipulation through the Spatial, Object, Goal, and Long suites.
The sample-efficiency study uses the same nested subsets of $N\in\{5,10,25,50\}$ demonstrations per task for every backbone on Goal and Long.
We additionally evaluate the full-data LIBERO policies, without retraining, under the environment and object perturbations of LIBERO-PRO~\cite{liberopro}.
RoboCasa~\cite{robocasa} tests kitchen manipulation in out-of-distribution scenes with a Panda gripper, using 30 or 100 generated demonstrations per task (Gen-30 and Gen-100).
From DexJoCo~\cite{dexjoco} we use the five bimanual dexterous tasks (Unlock iPad, Hanoi, Assembly, Microwave, and Photograph), with 100 human demonstrations per task.
We report object and full randomization separately and, for each setting, co-train one policy on all five tasks.
Example simulation tasks are shown in \cref{fig:tasks}. Evaluation settings are listed in \cref{sec:appendix_eval}.

\paragraph{VLM backbones.}
We compare DeepSeek-V4.1-Flash-1017 with Qwen3-VL and Cosmos-Reason2 backbones~\cite{qwen3vl,cosmosreason2}, listed in \cref{tab:backbones}.
Cosmos-Reason2-2B, used by GR00T N1.7~\cite{gr00tn17}, is the main small-backbone reference.
Within each benchmark, we keep the action DiT, its initialization, and the training data fixed across backbones.
Only the input width of the feature adapter follows the backbone.
No pretrained robot-action expert is loaded in these comparisons.

\begin{table}[t]
\centering
\caption{\textbf{VLM backbones used in the experiments.} Sizes refer to the total parameter count of each source model. MoE denotes a mixture of experts.}
\label{tab:backbones}
\footnotesize
\setlength{\tabcolsep}{3pt}
\begin{tabular}{lcc}
\toprule
\tabheader Backbone & Size label & Architecture \\
\midrule
Qwen3-VL-2B & 2B & Dense \\
Cosmos-Reason2-2B & 2B & Dense \\
Cosmos-Reason2-8B & 8B & Dense \\
Qwen3-VL-32B & 32B & Dense \\
Cosmos-Reason2-32B & 32B & Dense \\
\tabours DeepSeek-V4.1-Flash-1017 & 552B & MoE \\
\bottomrule
\end{tabular}
\end{table}

\paragraph{Training.}
Optimization updates only the action-side components, using the benchmark-specific recipes in \cref{sec:appendix_implementation}.
For each benchmark, we use the same training budget and checkpoint selection rule across backbones unless a training variant is marked separately.
We evaluate policies from the same initial states where the benchmark provides them.

\subsection{Simulation Results}
\label{sec:exp_main}

\methodname had the highest average success rate on each simulation benchmark. The largest gain in average success was on LIBERO-PRO.

\begin{figure*}[t]
  \centering
  \includegraphics[width=\textwidth]{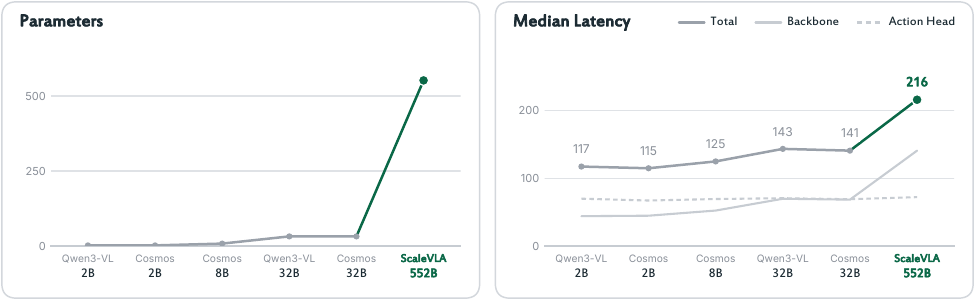}
  \caption{\textbf{Backbone size and inference latency.}
  Left: total parameters in each source model. Right: median time per action chunk on RoboCasa, with backbone and action head times shown separately.
  \methodname uses four H200 GPUs; the other backbones use one each.}
  \label{fig:latency}
\end{figure*}

\begin{table*}[t]
\centering
\caption{\textbf{LIBERO success rates (\%).} The first three rows report published results. Our policies train on each suite's full demonstration set. The last column averages the four suites. Bold marks the highest success rate in each column.}
\label{tab:libero}
\small
\setlength{\tabcolsep}{8pt}
\begin{tabular}{lccccc}
\toprule
\tabheader Model & Spatial & Object & Goal & Long & Average \\
\midrule
$\pi_0$~\cite{pi0} & 96.8 & 98.8 & 95.8 & 85.2 & 94.2 \\
OpenVLA~\cite{openvla} & 84.7 & 88.4 & 79.2 & 53.7 & 76.5 \\
CoT-VLA~\cite{cotvla} & 87.5 & 91.6 & 87.6 & 69.0 & 83.9 \\
\midrule
Qwen3-VL-2B & 98.4 & 98.2 & 90.4 & 77.8 & 91.2 \\
Cosmos-Reason2-2B & 98.8 & 97.6 & 83.2 & 83.2 & 90.7 \\
Cosmos-Reason2-8B & \textbf{99.8} & 99.0 & 88.6 & 87.2 & 93.7 \\
Qwen3-VL-32B & 98.8 & \textbf{99.4} & 93.2 & 83.0 & 93.6 \\
Cosmos-Reason2-32B & 98.4 & 98.4 & 95.2 & 83.8 & 94.0 \\
\tabours \methodname (552B) & 98.6 & 98.6 & \textbf{96.8} & \textbf{90.2} & \textbf{96.1} \\
\bottomrule
\end{tabular}
\end{table*}

\paragraph{LIBERO.}
Among our policies, success on Spatial and Object is near saturation. The larger differences are on Goal and Long (\cref{tab:libero}).
\methodname is the best backbone on both, with 96.8 on Goal and 90.2 on Long.
The four-suite average rises with backbone size in both dense families, from 90.7 for Cosmos-Reason2-2B and 91.2 for Qwen3-VL-2B to 94.0 and 93.6 at 32B, and reaches 96.1 for \methodname, above the three policies from prior work.
Goal requires grounding the instruction and Long requires chaining several steps, so the gains concentrate where success depends on understanding the task rather than on grasping alone.

\begin{table}[t]
\centering
\caption{\textbf{LIBERO-PRO success rates (\%).} We test our LIBERO policies from \cref{tab:libero} under environment (Env.) and object (Obj.) perturbations without retraining~\cite{liberopro}.}
\label{tab:libero_pro}
\footnotesize
\setlength{\tabcolsep}{4pt}
\begin{tabular}{lcccc}
\toprule
\tabheader & \multicolumn{2}{c}{Goal} & Object & \\
\cmidrule(lr){2-3}\cmidrule(lr){4-4}
\tabheader Backbone & Env. & Obj. & Env. & Mean \\
\midrule
Qwen3-VL-2B & 24.2 & 38.4 & 25.2 & 29.3 \\
Cosmos-Reason2-2B & 20.6 & 37.6 & 11.8 & 23.3 \\
Cosmos-Reason2-8B & 32.6 & 37.6 & 23.6 & 31.3 \\
Qwen3-VL-32B & 11.6 & 44.2 & 12.6 & 22.8 \\
Cosmos-Reason2-32B & 19.2 & 38.8 & \textbf{32.4} & 30.1 \\
\tabours \methodname (552B) & \textbf{47.8} & \textbf{56.4} & 31.2 & \textbf{45.1} \\
\bottomrule
\end{tabular}
\end{table}

\paragraph{LIBERO-PRO.}
Perturbations test whether a policy holds up outside its training conditions (\cref{tab:libero_pro}).
Every policy drops sharply, but \methodname retains the most: its mean of 45.1\% is about 14 points above the best dense backbone and well above the 23.3--29.3\% of the 2B backbones, and it leads under both Goal-suite perturbations.
Frontier representations therefore transfer better beyond the training distribution.
Environment perturbations of the Object suite remain hard for every backbone; the best entry is 32.4\%, and \methodname (31.2) is on par with it.

\begin{table}[t]
\centering
\caption{\textbf{RoboCasa success rates (\%).} Gen-30 and Gen-100 use 30 and 100 generated demonstrations per task, respectively. We test in out-of-distribution kitchens.}
\label{tab:robocasa}
\footnotesize
\setlength{\tabcolsep}{5pt}
\begin{tabular}{lcc}
\toprule
\tabheader Backbone & Gen-30 & Gen-100 \\
\midrule
Qwen3-VL-2B & 22.4 & 45.2 \\
Cosmos-Reason2-2B & 23.2 & 42.8 \\
Cosmos-Reason2-8B & 25.6 & 45.6 \\
Qwen3-VL-32B & 25.5 & 43.3 \\
\tabours \methodname (552B) & \textbf{30.7} & \textbf{54.9} \\
\bottomrule
\end{tabular}
\end{table}

\paragraph{RoboCasa.}
RoboCasa tests generalization to out-of-distribution kitchen scenes (\cref{tab:robocasa}).
\methodname is best with both 30 and 100 demonstrations per task.
The other backbones, from 2B to 32B, reach 22--26\% with 30 demonstrations and 43--46\% with 100, whereas \methodname reaches 30.7\% and 54.9\%.
The margin grows with the amount of data, from about five points to nine, so pretrained features and robot supervision complement each other.

\begin{table*}[t]
\centering
\caption{\textbf{DexJoCo success rates (\%).} We train one policy on all five bimanual tasks for each backbone and randomization setting (object or full). Bold marks the best of our backbones.}
\label{tab:dexjoco_cotrain}
\resizebox{\textwidth}{!}{%
\begin{tabular}{lcccccccccccc}
\toprule
\tabheader & DP-T~\cite{dp} & $\pi_{0.5}$~\cite{pi05} & \multicolumn{2}{c}{Qwen3-VL-2B} & \multicolumn{2}{c}{Cosmos-Reason2-2B} & \multicolumn{2}{c}{Cosmos-Reason2-8B} & \multicolumn{2}{c}{Qwen3-VL-32B} & \multicolumn{2}{c}{\methodname (552B)} \\
\cmidrule(lr){2-2}\cmidrule(lr){3-3}\cmidrule(lr){4-5}\cmidrule(lr){6-7}\cmidrule(lr){8-9}\cmidrule(lr){10-11}\cmidrule(lr){12-13}
\tabheader Task & Obj. & Obj. & Obj. & Full & Obj. & Full & Obj. & Full & Obj. & Full & Obj. & Full \\
\midrule
Unlock iPad & 0.0 & 0.0 & 17.3 & 8.7 & 15.3 & 4.7 & 26.7 & 6.0 & 26.7 & 14.0 & \textbf{31.3} & \textbf{15.3} \\
Hanoi & 8.0 & 4.7 & 11.3 & 8.7 & 24.0 & 4.7 & 24.0 & 13.3 & 24.0 & 10.7 & \textbf{28.0} & \textbf{14.0} \\
Assembly & 1.3 & 4.0 & 4.7 & 0.0 & 10.0 & 0.7 & 10.7 & 0.0 & 4.0 & 0.0 & \textbf{11.3} & \textbf{1.3} \\
Microwave & 42.7 & 34.7 & 84.7 & 61.3 & 81.3 & 62.7 & 82.0 & 62.0 & \textbf{89.3} & 62.7 & 73.3 & \textbf{67.3} \\
Photograph & 28.0 & 7.3 & 41.3 & \textbf{32.0} & 54.7 & 30.7 & 47.3 & 24.0 & 46.7 & 22.7 & \textbf{66.7} & 24.0 \\
\midrule
Avg. & 16.0 & 10.1 & 31.9 & 22.1 & 37.1 & 20.7 & 38.1 & 21.1 & 38.1 & 22.0 & \textbf{42.1} & \textbf{24.4} \\
\bottomrule
\end{tabular}}
\end{table*}


\begin{table*}[t]
\centering
\caption{\textbf{RoboCasa inference latency.} Times are medians per action chunk on H200 GPUs. Bold marks the lowest latency in each timing column.}
\label{tab:latency_robocasa}
\footnotesize
\setlength{\tabcolsep}{7pt}
\begin{tabular}{lccrrr}
\toprule
\tabheader & & & \multicolumn{3}{c}{Median latency (ms)} \\
\cmidrule(lr){4-6}
\tabheader Backbone & GPUs & Tokens & Backbone & Action head & Total \\
\midrule
Qwen3-VL-2B & 1 & 212 & \textbf{44.0} & 69.8 & 117.2 \\
Cosmos-Reason2-2B & 1 & 212 & 44.7 & \textbf{67.1} & \textbf{114.7} \\
Cosmos-Reason2-8B & 1 & 212 & 52.3 & 69.3 & 124.8 \\
Qwen3-VL-32B & 1 & 212 & 69.7 & 70.5 & 143.3 \\
Cosmos-Reason2-32B & 1 & 212 & 68.4 & 69.0 & 140.7 \\
\tabours \methodname (552B) & 4 & 568 & 140.7 & 72.1 & 215.8 \\
\bottomrule
\end{tabular}
\end{table*}

\begin{table}[t]
\centering
\caption{\textbf{LIBERO success rates (\%) by demonstration count.} On Goal and Long, we vary the number of demonstrations per task ($N$) and keep the training budget fixed. The 2B, 8B, and 32B backbones are Cosmos-Reason2 models. Bold marks the best backbone at each demonstration count within each suite.}
\label{tab:lowdata}
\footnotesize
\setlength{\tabcolsep}{4pt}
\begin{tabular}{lrrrrrrrr}
\toprule
\tabheader & \multicolumn{4}{c}{Goal} & \multicolumn{4}{c}{Long} \\
\cmidrule(lr){2-5}\cmidrule(lr){6-9}
\tabheader $N$ & 2B & 8B & 32B & 552B & 2B & 8B & 32B & 552B \\
\midrule
5 & 39.4 & 60.6 & 61.2 & \textbf{77.4} & 24.0 & 40.0 & 37.6 & \textbf{62.2} \\
10 & 55.0 & 70.0 & 83.0 & \textbf{87.8} & 56.0 & 65.2 & 66.4 & \textbf{75.0} \\
25 & 85.2 & 83.0 & 90.4 & \textbf{97.6} & 69.6 & 65.6 & 83.6 & \textbf{84.6} \\
50 & 85.6 & 88.6 & 95.2 & \textbf{95.4} & 82.4 & \textbf{87.2} & 83.8 & 86.8 \\
\bottomrule
\end{tabular}
\end{table}

\paragraph{DexJoCo.}
\methodname had the highest average success rate on the five bimanual tasks under both object and full randomization (\cref{tab:dexjoco_cotrain}). With object randomization, average success rose from 31.9--37.1\% at 2B to 38.1\% at 8B and 32B, then to 42.1\% for \methodname. Its advantage was most consistent on Unlock iPad, Hanoi, and Assembly, the three tasks with the lowest success rates among our policies. \methodname led on all three in both settings. With full randomization, for example, it reached 15.3\% on Unlock iPad and 14.0\% on Hanoi, compared with 8.7\% and 10.7\% for the next best backbone. Assembly remained difficult for all our policies, with at most 1.3\% success under full randomization.

Each policy learns all five tasks in one randomization setting, so performance depends on handling different tasks and layouts with the same weights. Full randomization reduced success for every backbone, but \methodname retained the highest average at 24.4\%, compared with 22.1\% for the best smaller backbone. Results on Microwave and Photograph varied by setting. The published DP-T and $\pi_{0.5}$ baselines average 16.0\% and 10.1\%, respectively~\cite{dp,pi05,dexjoco}. The comparison across our backbones shows the clearest advantage on the harder bimanual tasks that require precise coordination.

\subsection{Inference Latency}
\label{sec:exp_latency}

\methodname reads VLM features in one forward pass per action chunk. We measure the time from sending the observation to receiving the action chunk at the client, following \cref{sec:appendix_timing}.

\paragraph{Latency increases less than proportionally with parameter count.}
On RoboCasa, \methodname has 276$\times$ the parameters of Cosmos-Reason2-2B but 1.9$\times$ the median latency per action chunk (216 \vs 115\,ms; \cref{fig:latency,tab:latency_robocasa}).
The sparse mixture-of-experts backbone activates only a subset of its experts per token.
We run the causal encoder once per chunk and reuse its features during action generation.
With four H200 GPUs, this keeps latency in the hundreds of milliseconds per chunk.
On the real robot, \methodname ran the Plug In Charger task at a median round-trip latency of 384\,ms per chunk (\cref{tab:real}).

\paragraph{Action head runtime.}
The action head takes about 70\,ms for every backbone, which is more than half of the total for a 2B backbone.
Growing the backbone from 2B to 32B therefore adds only about 26\,ms, and larger backbones add to this fixed cost instead of multiplying it.

\paragraph{GPU requirements.}
\methodname uses four GPUs with tensor and expert parallelism; the smaller backbones use one GPU each.
\methodname also receives more input tokens per observation. We did not isolate how much this contributes to latency.

\subsection{Sample Efficiency}
\label{sec:exp_lowdata}

We vary the number of demonstrations per task, $N\in\{5,10,25,50\}$, on LIBERO Goal and Long while fixing training at 5,000 optimizer steps.
All backbones use the same nested demonstration subsets and evaluation initial states.

\paragraph{Sample efficiency across backbone scales.}
As shown in \cref{tab:lowdata}, \methodname consistently
outperforms smaller backbones in low-data regimes, with
the most pronounced gains when demonstrations are scarce.
With only five demonstrations per task, scaling the backbone
from 2B to 552B nearly doubles the success rate on
LIBERO-Goal, from 39.4\% to 77.4\%, while yielding substantial
improvements on LIBERO-Long. Notably, \methodname requires
only one-fifth of the demonstrations to surpass the 2B
baseline on LIBERO-Goal. These results suggest that
frontier-scale pretrained representations provide a strong
foundation for action learning, enabling effective policy
learning from limited robot demonstrations.


\paragraph{Interplay between backbone scale and demonstration availability.}
As the number of demonstrations increases, the performance gap 
between backbones generally narrows, suggesting that richer 
pretrained representations are particularly beneficial in 
data-scarce regimes. While \methodname approaches saturation 
on LIBERO-Goal with only 25 demonstrations per task, its 
performance on LIBERO-Long continues to benefit from 
additional supervision. These results suggest that scaling 
pretrained representations reduces the reliance on robot 
demonstrations, while additional task-specific experience 
remains valuable for long-horizon manipulation.


\begin{figure*}[t]
  \centering
  \includegraphics[width=\textwidth]{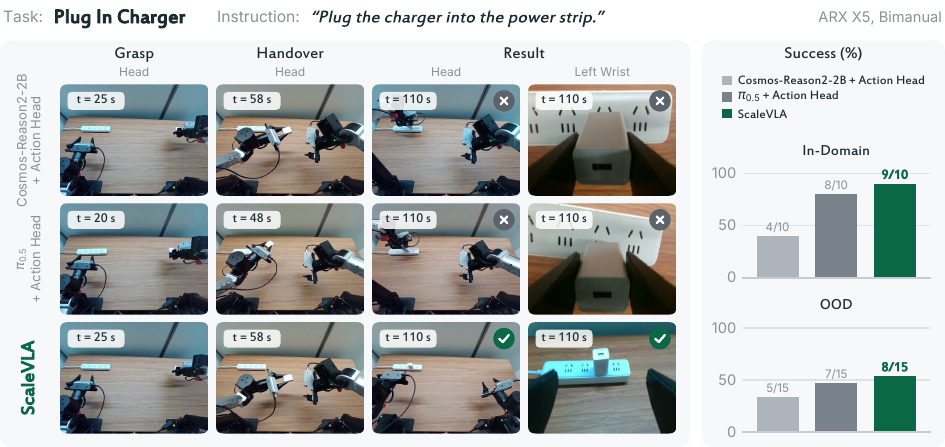}
  \caption{\textbf{Plug In Charger on a real robot.}
  Left: all three policies grasp and hand over the charger from in-domain layouts. \methodname aligns and inserts the plug; the other two policies reach the socket but fail to insert the plug.
  Right: success rates in the in-domain and OOD settings.}
  \label{fig:real}
\end{figure*}

\begin{table*}[t]
\centering
\caption{\textbf{Real-robot success rates and timing.} Successes are counted over ten in-domain and fifteen out-of-distribution (OOD) trials. Insertion and completion times are medians over successful in-domain trials. Latency is the median round trip per action chunk at the robot client. Bold marks the most successes or the shortest time.}
\label{tab:real}
\footnotesize
\setlength{\tabcolsep}{4pt}
\begin{tabular}{lccccc}
\toprule
\tabheader & \multicolumn{2}{c}{Successes} & \multicolumn{2}{c}{In-domain, median (s)} & Latency \\
\cmidrule(lr){2-3}\cmidrule(lr){4-5}
\tabheader Policy & In-domain & OOD & Insertion & Completion & (ms) \\
\midrule
Cosmos-Reason2-2B + Action Head & 4/10 & 5/15 & 85.9 & 147.1 & 311 \\
$\pi_{0.5}$ + Action Head & 8/10 & 7/15 & 81.2 & 132.0 & \textbf{262} \\
\tabours \methodname (552B) & \textbf{9/10} & \textbf{8/15} & \textbf{58.1} & \textbf{114.4} & 384 \\
\bottomrule
\end{tabular}
\end{table*}

\subsection{Real-Robot Experiments}

\paragraph{Overview.}
To investigate whether the benefits of scaling VLM backbones 
extend beyond simulation, we conduct real-world experiments 
on a bimanual robot. We use the Plug In Charger task 
from RoboDojo~\cite{chen2026robodojo}, which in our setup requires sequential execution of
object grasping, bimanual handover, and precise, contact-rich 
insertion. This task provides a challenging testbed for 
evaluating whether frontier-scale visual representations 
can benefit both multi-stage manipulation and fine-grained 
robotic control. We further evaluate policy robustness under 
spatial perturbations and examine the practical feasibility 
of deploying a frontier-scale VLA policy on physical robots.

\label{sec:real}

\paragraph{Training data.}
The task requires the right arm to grasp a phone charger and hand it over, and the left arm to align the plug with a power-strip socket, insert it, and release it, observed through a top camera and two wrist cameras.
We train on full teleoperated demonstrations and additional segments focused on insertion. Most demonstrations start near one reference charger position.

\paragraph{Models.}
We compare three backbones under the same action DiT, training data, and training budget: Cosmos-Reason2-2B, $\pi_{0.5}$~\cite{pi05}, and \methodname. Cosmos-Reason2-2B and \methodname have similar validation action errors (\cref{sec:appendix_eval}). Unlike the other two backbones, the VLM of $\pi_{0.5}$ was trained on large-scale robot data as part of $\pi_{0.5}$, so it tests whether robot pretraining of a smaller backbone can substitute for frontier-scale representations.

\paragraph{Real-robot evaluation protocol.}
A trial succeeds when the plug is inserted and the left gripper releases it.
Each policy runs ten in-domain trials, which start near the dominant training position, and fifteen OOD trials, in which we shift or rotate the charger or move the power strip beyond most training placements.

\paragraph{Success rates.}
As shown in \cref{tab:real}, \methodname achieves the highest success rate under both in-domain and OOD settings. Robot pretraining narrows but does not close the gap: $\pi_{0.5}$ + Action Head succeeds in 8 of 10 in-domain and 7 of 15 OOD trials, compared with 9 of 10 and 8 of 15 for \methodname, and needs a median 81.2\,s rather than 58.1\,s to insert the plug. Notably, the benefits of frontier-scale representations are most evident in the contact-rich insertion stage. While all three policies exhibit comparable execution times during grasping and handover in successful trials, they differ substantially in their ability to align and insert the charger. All in-domain failures occur during insertion, whereas \methodname completes this stage approximately 30\% faster than the smaller-backbone policies. This suggests that scaling the VLM backbone can benefit not only high-level task understanding but also fine-grained manipulation that requires precise spatial perception and control.

Under spatial distribution shifts, \methodname maintains the highest success rate, although precise insertion outside the training distribution remains challenging. Interestingly, while the Cosmos-Reason2-2B policy encounters multiple safety stops due to insufficient arm clearance, \methodname exhibits no such failures across all trials. Together, these observations suggest that frontier-scale pretrained representations can support more reliable visuomotor control in physically demanding manipulation tasks.

\paragraph{Completion time and latency.}
Despite the increased inference latency of its frontier-scale backbone, \methodname achieves the shortest end-to-end completion time among all evaluated policies (\cref{tab:real}). This advantage primarily comes from more efficient execution of the contact-rich insertion stage. Interestingly, these results reveal that faster policy inference does not necessarily translate into faster task execution. For manipulation tasks requiring precise spatial alignment, improvements in action execution efficiency can outweigh the additional computational cost of scaling the backbone. This highlights the importance of evaluating the practical efficiency of VLA policies at the task level rather than solely through latency.

\section{Conclusion}
\label{sec:conclusion}

We introduced \methodname, which grounds the representations of a 552B VLM in continuous robot actions through an action DiT trained with flow matching.
Across LIBERO, LIBERO-PRO, RoboCasa, and DexJoCo, performance rises with backbone scale, and on a real robot the 552B backbone inserts a charger more reliably than the 2B backbone.
Frontier features prove most valuable where robot data is scarce, conditions shift, or contact must be precise, and they remain fast enough for subsecond control, although on more hardware than smaller backbones.
These results suggest that frontier VLMs can serve not only as reasoning agents, but also as the representational foundation of fast, learned robot policies.

\bibliographystyle{assets/plainnat}
\bibliography{main}

\appendix
\clearpage
\section{Implementation Details}
\label{sec:appendix_implementation}

\paragraph{Feature extraction.}
DeepSeek-V4.1-Flash-1017 supplies 5120-dimensional visual and text features at the causal encoder--decoder interface described in \cref{sec:backbone}.
We retain the token sequence and modality labels, and cache the features for training.
The smaller-backbone experiments use intermediate token representations with a readout depth matched by a common relative-depth rule.
The previous comparison protocol uses layer $\operatorname{round}(16L/28)$ for an $L$-layer language model, giving readouts 16/28, 21/36, and 37/64 for the recorded 2B, 8B, and 32B backbones.

\paragraph{Action head.}
The backbone features pass through layer normalization, two linear projections to a width of 2048, and four layers of self-attention.
The action DiT has 32 layers and alternates self-attention with cross-attention to the adapted features.
State and noisy-action encoders, flow-time conditioning, and the embodiment-specific velocity decoder follow \cref{sec:action_head}.
These action-side components are randomly initialized; no pretrained robot-action expert is loaded.
The action DiT follows the GR00T N1.7 architecture~\cite{gr00tn17}.

\paragraph{Optimization and normalization.}
Policies are trained on 64 NVIDIA H200 GPUs (eight nodes with eight GPUs each), with a global batch size of 640 and policy-training seed 42.
For LIBERO, the initial full-data recipe uses 5,000 AdamW steps at learning rate $10^{-4}$ with cosine decay and 5\% warmup.
The sample-efficiency experiment holds this training budget fixed and varies demonstrations per task (\cref{tab:lowdata}); a separate study varies update count at five demonstrations per task (\cref{tab:lowdata_steps}).
DexJoCo uses 20,000 optimizer steps at learning rate $10^{-4}$, with cosine decay and 5\% warmup.
State and action normalization uses the 1st and 99th percentiles of the corresponding training data.
In the demonstration-budget sweep, policy updates use the selected subsets, but normalization retains statistics from the full training dataset.
Thus, the reported demonstration counts describe trajectories used for policy updates, not all information used in preprocessing.

\section{Execution and Latency Measurement}
\label{sec:appendix_timing}

LIBERO predicts 16-step chunks and executes eight commands before obtaining the next policy decision.
DexJoCo predicts 30-step chunks, assigns commands to execution timestamps, and computes replacement chunks while the controller consumes the buffer.
Both run four flow steps using the backbone features from one forward pass per chunk.
For each deployment setting, the timing report should specify hardware allocation, batch size, warmup, the number of timed calls, and whether other jobs run concurrently on the machine.
Preprocessing, backbone inference, feature transfer, action-head inference, and postprocessing should be measured together for end-to-end chunk latency.

\paragraph{Asynchronous execution.}
DexJoCo can continue executing buffered commands while the policy generates the next chunk. These commands may use older observations, so controller frequency alone does not measure how quickly the policy reacts to new feedback. To characterize the feedback loop, report chunk latency, observation age at execution, and intervals when the buffer has no valid commands.

\paragraph{RoboCasa timing.}
\Cref{tab:latency_robocasa} and \cref{fig:latency} were measured on one machine with eight H200 GPUs, one request at a time with batch size one.
After 50 warm-up calls, each backbone handles 3,000 timed calls, made of 1,000 recorded RoboCasa observations (three camera images, robot state, and instruction) replayed three times.
The observations come from rollouts of a Cosmos-Reason2-2B policy on eight RoboCasa tasks.
A call is timed at the client, from sending the observation to receiving the action chunk. Client-side latency adds about 3\,ms to the server time. The token counts in \cref{tab:latency_robocasa} are typical input lengths per call. The measured median for \methodname corresponds to 4.6 chunks per second for sequential requests.
The server splits this time into backbone time, which includes message passing and preprocessing, and action-head time, which covers the feature adapter, four flow steps, and state and action processing.
The action head predicts eight steps per chunk and runs without CUDA graphs on a GPU that the backbone already uses.

All backbones are served with vLLM 0.31 and the same settings: batch-invariant kernels, eager execution, and no prefix cache.
Each model is cut at the readout layer used in training and runs one prefill pass per call.
\methodname uses four GPUs with tensor and expert parallelism, and the other backbones use one GPU.
Each backbone uses its own RoboCasa Gen-100 checkpoint.
vLLM serves for timing only: its features are not compared element by element with those of the Hugging Face deployment that produced the success rates.
For every backbone except \methodname, the prompt token ids and types of the two paths match on 16 observations.
The single-GPU backbones were timed in four parallel lanes, one GPU each, while a separate RoboCasa evaluation ran on the other four GPUs.
\methodname was timed in an earlier run with the same settings.

\section{Evaluation Protocols}
\label{sec:appendix_eval}

\paragraph{LIBERO and LIBERO-PRO.}
Our LIBERO evaluation uses ten tasks per suite and 50 official initial states per task. LIBERO-PRO uses the official task instructions and 500 episodes per setting.
The overall scores are unweighted means over the four LIBERO suites and three LIBERO-PRO settings, respectively.

\paragraph{RoboCasa.}
Each success rate averages 24 tasks, with 50 episodes per task and a 720-step limit.
Gen-30 uses the better result from checkpoints at 48,000 and 96,000 steps; Gen-100 uses the checkpoint at 96,000 steps.

\paragraph{DexJoCo.}
We use the official evaluator with three evaluation seeds and 50 episodes per task and seed.
The overall score is the unweighted mean of the five task success rates.
Published baselines are taken directly from Table IV of \citet{dexjoco}.

\paragraph{Real-robot evaluation.}
Training uses 171 full teleoperated demonstrations and 132 insertion segments (303 episodes, 0.52 million frames); 26 episodes are kept for validation.
Most full demonstrations place the charger within a few centimeters of one reference position, with a minority shifted by up to about 13\,cm.
After 40,000 training steps, the Cosmos-Reason2-2B and \methodname policies both reach a training loss of 0.0016. Validation mean absolute action errors are 0.0095 for the 2B policy and 0.0094 for \methodname.
$\pi_{0.5}$ + Action Head reads features from layer 10 of the PaliGemma backbone of $\pi_{0.5}$~\cite{pi05} and trains the same action DiT on the same data for 40,000 steps.
All policies execute 16-step action chunks with identical safety limits on the same client.

Each policy runs ten in-domain trials and fifteen out-of-distribution (OOD) trials.
In the in-domain trials, the charger is within a few centimeters of the dominant training position.
OOD layouts shift the charger up to about 16\,cm toward the robot or the workspace center, rotate it by about $30^\circ$ in either direction, or move the power strip by about 10\,cm or to the back edge of the table.
The operator judges success. Times are wall-clock times from the client log and include inference: completion runs from the first action request to the last action chunk, and insertion runs from the start of the insertion stage, as logged by the client, to the last action chunk.
Latency is the round trip per action chunk measured at the robot client, from sending the observation to receiving the chunk, with the median taken over all chunks of all trials of a policy; this is the same measurement as the RoboCasa latency in \cref{tab:latency_robocasa}.

Each policy we train uses one training seed. Repeating evaluation episodes measures performance across test conditions.

\section{Additional Results}
\label{sec:appendix_results}

\subsection{Training Steps with Five Demonstrations}
\label{sec:appendix_lowdata}

\begin{table}[H]
\centering
\caption{\textbf{LIBERO success rates (\%) by training budget.} Cosmos-Reason2-2B and \methodname each train with five demonstrations per task. Bold marks the higher success rate when both backbones have results. A dash indicates an unreported result.}
\label{tab:lowdata_steps}
\footnotesize
\setlength{\tabcolsep}{3pt}
\begin{tabular}{llrrrrr}
\toprule
\tabheader Suite & Backbone & 2k & 5k & 10k & 20k & 40k \\
\midrule
Goal & 2B & 33.4 & 39.4 & 51.4 & 67.4 & 69.6 \\
\tabours     & 552B & \textbf{77.8} & \textbf{77.4} & \textbf{76.4} & \textbf{77.2} & -- \\
\midrule
Long & 2B & 10.4 & 24.0 & 39.0 & 50.6 & 55.2 \\
\tabours     & 552B & \textbf{46.4} & \textbf{62.2} & \textbf{67.0} & \textbf{71.8} & -- \\
\bottomrule
\end{tabular}
\end{table}

\Cref{tab:lowdata_steps} varies optimizer updates while retaining five demonstrations per task.
On Goal, the frontier policy reaches a stable level early; on Long, it continues improving with additional updates.
The small-backbone policy benefits from longer training on both suites, so fixed-budget sample efficiency should not be equated with a comparison after convergence.
The 5k column is the $N=5$ row of \cref{tab:lowdata}.

\paragraph{Demonstration sweep.}
The values for Cosmos-Reason2-2B and \methodname at $N=50$ in \cref{tab:lowdata} come from the demonstration sweep rather than the full-data evaluation in \cref{tab:libero}.
Both evaluations use 50 official initial states per task and a 720-step limit.
The sweep uses one sequence of nested demonstration subsets.

\end{document}